# Japanese-Spanish Thesaurus Construction Using English as a Pivot

**Jessica Ramírez, Masayuki Asahara, Yuji Matsumoto**
Graduate School of Information Science
Nara Institute of Science and Technology
Ikoma, Nara, 630-0192 Japan
`{jessic-r,masayu-a,matsu}@is.naist.jp`

## Abstract

We present the results of research with the goal of automatically creating a multilingual thesaurus based on the freely available resources of Wikipedia and WordNet. Our goal is to increase resources for natural language processing tasks such as machine translation targeting the Japanese-Spanish language pair. Given the scarcity of resources, we use existing English resources as a pivot for creating a trilingual Japanese-Spanish-English thesaurus. Our approach consists of extracting the translation tuples from Wikipedia, disambiguating them by mapping them to WordNet word senses. We present results comparing two methods of disambiguation, the first using VSM on Wikipedia article texts and WordNet definitions, and the second using categorical information extracted from Wikipedia, We find that mixing the two methods produces favorable results. Using the proposed method, we have constructed a multilingual Spanish-Japanese-English thesaurus consisting of 25,375 entries. The same method can be applied to any pair of languages that are linked to English in Wikipedia.

## 1 Introduction

Aligned data resources are indispensable components of many Natural Language Processing (NLP) applications; however lack of annotated data is the main obstacle for achieving high performance NLP systems. Current success has been moderate. This is because for some languages there are few resources that are usable for NLP.

Manual construction of resources is expensive and time consuming. For this reason NLP researchers have proposed semi-automatic or automatic methods for constructing resources such as dictionaries, thesauri, and ontologies, in order to facilitate NLP tasks such as word sense disambiguation, machine translation and other tasks. Hoglan Jin and Kam-Fai Wong (2002) automatically construct a Chinese dictionary from different Chinese corpora, and Ahmad Khurshid et al. (2004) automatically develop a thesaurus for a specific domain by using text that is related to an image collection to aid in image retrieval.

With the proliferation of the Internet and the immense amount of data available on it, a number of researchers have proposed using the World Wide Web as a large-scale corpus (Rigau et al., 2002). However due to the amount of redundant and ambiguous information on the web, we must find methods of extracting only the information that is useful for a given task.

### 1.1 Goals

This research deals with the problem of developing a multilingual Japanese-English-Spanish thesaurus that will be useful to future Japanese-Spanish NLP research projects.

A thesaurus generally means a list of words grouped by concepts; the resource that we create is similar because we group the words according to semantic relations. However, our resource is also composed of three languages – Spanish, English, and Japanese. Thus we call the resource we created a multilingual thesaurus.

Our long term goal is the construction of a Japanese-Spanish MT system. This thesaurus will be used for word alignments and building comparable corpus.

We construct our multilingual thesaurus by following these steps:
- Extract the translation tuples from Wikipedia article titles
- Align the word senses of these tuples with those of English WordNet (disambiguation)
- Construct a parallel thesaurus of Spanish-English-Japanese from these tuples

### 1.2 Method summary

We extract the translation tuples using Wikipedia's hyperlinks to articles in different languages and align these tuples to WordNet by measuring cosine vector similarity measures between Wikipedia article texts and WordNet glosses. We also use heuristics comparing the Wikipedia categories of a word with its hypernyms in WordNet.

A fundamental step in the construction of a thesaurus is part of speech (POS) identification of words and word sense disambiguation (WSD) of polysemous entries.

For POS identification, we cannot use Wikipedia, because it does not contain POS information. So we use another well-structured resource, WordNet, to provide us with the correct POS for a word.

These two resources, Wikipedia and WordNet, contain polysemous entries. We also introduce WSD method to align these entries.

We focus on the multilingual application of Wikipedia to help transfer information across languages. This paper is restricted mainly to nouns, noun phrases, and to a lesser degree, named entities, because we only use Wikipedia article titles.

## 2 Resources

### 2.1 Wikipedia

Wikipedia is an online multilingual encyclopedia with articles on a wide range of topics, in which the texts are aligned across different languages.

Wikipedia has some features that make it suitable for research such as:

Each article has a title, with a unique ID. "Redirect pages" handle synonyms, and "disambiguation pages" are used when a word has several senses. "Category pages" contain a list of words that share the same semantic category. For example the category page for "Birds" contains links to articles like "parrot", "penguin", etc. Categories are assigned manually by users and therefore not all pages have a category label.

Some articles belong to multiple categories. For example, the article "Dominican Republic" belongs to three categories: "Dominican Republic", "Island countries" and "Spanish-speaking countries". Thus, the article Dominican Republic appears in three different category pages.

The information in redirect pages, disambiguation pages and Category pages combines to form a kind of Wikipedia taxonomy, where entries are identified by semantic category and word sense.

### 2.2 WordNet

WordNet (C. Fellbaum, 1998) *"is considered to be one of the most important resources in computational linguistics and is a lexical database, in which concepts have been grouped into sets of synonyms (words with different spellings, but the same meaning), called synsets, recording different semantic relations between words"*.

WordNet can be considered to be a kind of machine-readable dictionary. The main difference between WordNet and conventional dictionaries is that WordNet groups the concepts into *synsets*, and each concept has a small definition sentence call a "gloss" with one or more sample sentences for each *synset*.

When we look for a word in WordNet it presents a finite number of synsets, each one representing a concept or idea.

The entries in WordNet have been classified according to the syntactic category such as: nouns, verbs, adjectives and adverbs, etc. These syntactic categories are known as part of speech (POS).

## 3 Related Work

Compared to well-established resources such as WordNet, there are currently comparatively fewer researchers using Wikipedia as a data resource in NLP. There are, however, works showing promising results.

The work most closely related to this paper is (M. Ruiz et al., 2005), which attempts to create an ontology by associating the English Wikipedia links with English WordNet. They use the "Simple English Wikipedia" and WordNet version 1.7 to measure similarity between concepts. They compared the WordNet glosses and Wikipedia by using the Vector Space Model, and presented results using the cosine similarity.

Our approach differs in that we disambiguate the Wikipedia category tree using WordNet hyper-/hyponym tree. We compare our approach to M. Ruiz et al., (2005) using it as the baseline in section 7.

Oi Yee Kwong (1998) integrates different resources to construct a thesaurus by using WordNet as a pivot to fill gaps between thesaurus and a dictionary.

Strube and Ponzetto (2006) present some experiments using Wikipedia for the computing semantic relatedness of words (a measure of degree to which two concepts are related in a taxonomy measured using all semantic relations), and compare the results with WordNet. They also integrate Google hits, in addition to Wikipedia and WordNet based measures.

## 4 General Description

First we extract from Wikipedia all the aligned links i.e. Wikipedia article titles. We map these on to WordNet to determine if a word has more than one sense (polysemous) and extract the ambiguous articles. We use two methods to disambiguate by assigning the WordNet sense to the polysemous words, we use two methods:

- Measure the cosine similarity between each Wikipedia article's content and the WordNet glosses.
- Compare the Wikipedia category to which the article belongs with the corresponding word in WordNet's ontology

Finally, we substitute the target word into Japanese and Spanish.

## 5 Extracting links from Wikipedia

The goal is the acquisition of Japanese-Spanish-English tuples of Wikipedia's article titles. Each Wikipedia article provides links to corresponding articles in different languages.

Every article page in Wikipedia has on the left hand side some boxes labeled: 'navigation', 'search', 'toolbox' and finally 'in other languages'. This has a list of all the languages available for that article, although the articles in each language do not all have exactly the same contents. In most cases English articles are longer or have more information than their counterparts in other languages, because the majority of Wikipedia collaborators are native English speakers.

**Pre-processing procedure:**

Before starting with the above phases, we first eliminate the irrelevant information from Wikipedia articles, to make processing easy and faster. The steps applied are as follows:

1. Extract the Wikipedia web articles
2. Remove from the pages all irrelevant information, such as images, menus, and special markup such as: "()", """, "*", etc...
3. Verify if a link is a redirected article and extract the original article
4. Remove all stopwords and function words that do not give information about a specific topic such as "the", "between", "on", etc.

**Methodology**

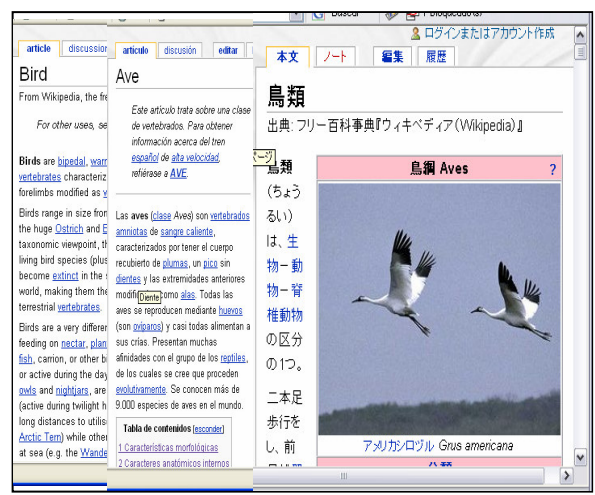

Figure 1. The article "bird" in English, Spanish and Japanese

Take all articles titles that are nouns or named entities and look in the articles' contents for the box called '*In other languages*'. Verify that it has at least one link. If the box exists, it links to the same article in other languages. Extract the titles in these other languages and align them with the original article title.

For instance, Figure 1. shows the English article titled "bird" translated into Spanish as "ave", and into Japanese as "chourui" (鳥類). When we click Spanish or Japanese '*in other languages*' box, we obtain an article about the same topic in the other language. This gives us the translation as its title, and we proceed to extract it.

## 6 Aligning Wikipedia entries to WordNet senses

The goal of aligning English Wikipedia entries to WordNet 2.1 senses is to disambiguate the polysemous words in Wikipedia by means of comparison with each sense of a given word existing in WordNet.

A gloss in WordNet contains both an association of POS and word sense. For example, the entry "bark#n#1" is different than "bark#v#1" because their POSes are different. In this example, "n" denotes noun and "v" denotes verb. So when we align a Wikipedia article to a WordNet gloss, we obtain both POS and word sense information.

**Methodology**

We assign WordNet senses to Wikipedia's polysemous articles. Firstly, after extracting all links and their corresponding translations in Spanish and Japanese, we look up the English words in WordNet and count the number of senses that each word has. If the word has more than one sense, the word is polysemous.

We use two methods to disambiguate the ambiguous articles, the first uses cosine similarity and the second uses Wikipedia's category tree and WordNet's ontology tree.

### 6.1 Disambiguation using Vector Space Model

We use a Vector Space Model (VSM) on Wikipedia and WordNet to disambiguate the POS and word sense of Wikipedia article titles. This gives us a correspondence to a WordNet gloss.

$$\cos \theta = \frac{V_1 \cdot V_2}{|V_1| \cdot |V_2|}$$

Where $V_1$ represents the Wikipedia article's word vector and $V_2$ represents the WordNet gloss' word vector.

In order to transfer the POS and word sense information, we have to measure similarity metric between a Wikipedia article and a WordNet gloss.

**Background**

VSM is an algebraic model, in which we convert a Wikipedia article into a vector and compares it to a WordNet gloss (that has also been converted into a vector) using the cosine similarity measure. It takes the set of words in some Wikipedia article and compares them with the set of words of WordNet gloss. Wikipedia articles which have more words in common are considered similar documents.

In Figure 2 shows the vector of the word "bank", we want to compare the similitude between the Wikipedia article "bank-1" with the English WordNet "bank-1" and "bank-2".

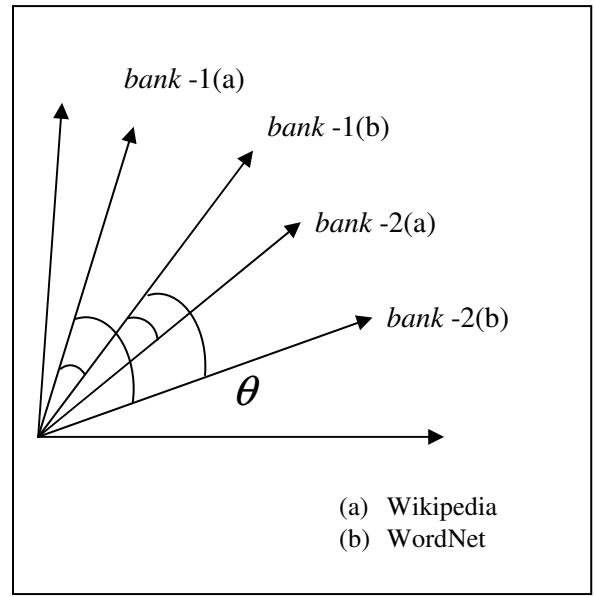

Figure 2. Vector Space Model with the word "bank"

VSM Algorithm:

1. Encode the Wikipedia article as a vector, where each dimension represents a word in the text of the article
2. Encode the WordNet gloss of each sense as a vector in the same manner

3. Compute the similarity between the Wikipedia vector and WordNet senses' vectors for a given word using the cosine measure
4. Link the Wikipedia article to the WordNet gloss with the highest similarity

## 6.2 Disambiguation by mapping the WordNet ontological tree to Wikipedia categories

This method consists of mapping the Wikipedia Category tree to the WordNet ontological tree, by comparing hypernyms and hyponyms. The main assumption is that there should be overlap between the hypernyms and hyponyms of Wikipedia articles and their correct WordNet senses. We will refer to this method as MCAT ("Map CATegories") throughout the rest of this paper.

Wikipedia has in the bottom of each page a box containing the category or categories to which the page belongs, as we can see in Figure 3. Each category links to the corresponding category page to which the title is affiliated. This means that the "category page" contains a list of all articles that share a common category.

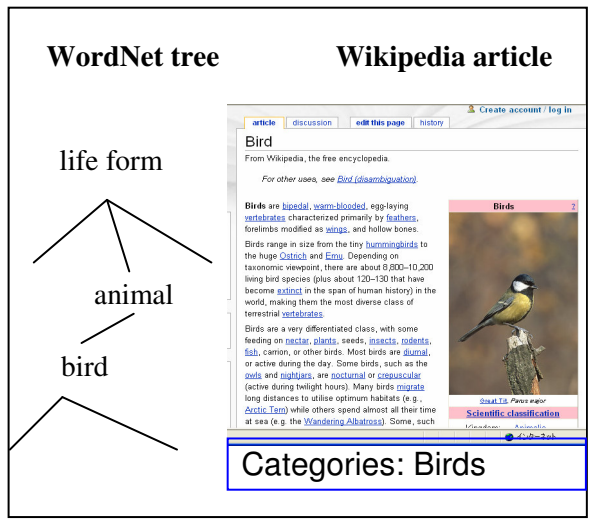

Figure 3. Relation between WordNet ontological tree and Wikipedia categories

**Methodology**

1. We extract ambiguous Wikipedia article titles (links) and the corresponding category pages
2. Extract the category pages, containing all pages which belong to that category, its subcategories, and other category pages that have a branch in the tree and categories to which it belongs.
3. If the page has a category:
3.1 Construct an n-dimensional vector containing the links and their categories
3.2 Construct an n-dimensional vector of the category pages, where every dimension represents a link which belongs to that category
4. For each category that an article belongs to:
4.1 Map the category to the WordNet hypernym-/hyponym tree by looking in each place that the given word appears and verify if any of its branches exist in the category page vector.
4.2 If a relation cannot be found then continue with other categories
4.3 If there is no correspondence at all then take the category pages vector and look to see if any of the links has relation with the WordNet tree
5. If there is at least one correspondence then assign this sense

## 6.3 Constructing the multilingual thesaurus

After we have obtained the English words with its corresponding English WordNet sense aligned in the three languages, we construct a thesaurus from these alignments.

The thesaurus contains a unique ID for every tuple of word and POS that it will have information about the syntactic category.

It also contains the sense of the word (obtain in the disambiguation process) and finally a small definition, which have the meaning of the word in the three languages.
- We assign a unique ID to every tuple of words
- For Spanish and Japanese we assign for default sense 1 to the first occurrence of the word if there exists more than 1 occurrence we continue incrementing
- Extract a small definition from the corresponding Wikipedia articles

The definition of title word in Wikipedia tends to be in the first sentence of the article.

Wikipedia articles often include sentences defining the meaning of the article's title. We mine Wikipedia for these sentences include them in our thesaurus. There is a large body of research dedicate to identifying definition sentences (Wilks et al., 1997), However, we currently rely on very simple patterns to this (e.g. "X is a/are Y", "X es un/a Y", "X は/が Y である。"). Incorporating more sophisticated methods remains an area of future work.

# 7 Experiments

## 7.1 Extracting links from Wikipedia

We use the articles titles from Wikipedia which are mostly nouns (including named entities) in Spanish, English and Japanese; (es.wikipedia.org, en.wikipedia.org, and ja.wikipedia.org), specifically "the latest all titles" and "the latest pages articles" files retrieved in April of 2006, and English WordNet version 2.1.

Our Wikipedia data contains a total of 377,621 articles in Japanese; 2,749,310 in English; and 194,708 in Spanish. We got a total of 25,379 words aligned in the three languages.

## 7.2 Aligning Wikipedia entries to WordNet senses

In WordNet there are 117,097 words and 141,274 senses. In Wikipedia (English) there are 2,749,310 article titles. 78,247 word types exist in WordNet. There are 14,614 polysemous word types that will align with one of the 141,274 senses in WordNet.

We conduct our experiments using 12,906 ambiguous articles from Wikipedia.

Table 1 shows the results obtained for WSD. The first column is the baseline (M. Ruiz et al., 2005) using the whole article; the second column is the baseline using only the first part of the article.

The third column (MCAT) shows the results of the second disambiguation method (disambiguation by mapping the WordNet ontological tree to Wikipedia categories). Finally the last column shows the results of combined method of taking the MCAT results when available and falling back to MCAT otherwise. The first row shows the sense assignments, the second row shows the incorrect sense assignment, and the last row shows the number of word used for testing.

### 7.2.1 Disambiguation using VSM

In the experiment using VSM, we used human evaluation over a sample of 507 words to verify if a given Wikipedia article corresponds to a given WordNet gloss. We took a the stratified sample of our data selecting the first 5 out of every 66 entries as ordered alphabetically for a total of 507 entries.

We evaluate the effectiveness of using whole articles in Wikipedia versus only a part (the first part up to the first subtitle), we found that the best score was obtained when using the whole articles 81.5% (410 words) of them are correctly assigned and 18.5% (97 words) incorrect.

**Discussion**

In this experiment because we used VSM the result was strongly affected by the length of the glosses in WordNet, especially in the case of related definitions because the longer the gloss the greater the probability of it having more words in common.

An example of related definitions in English WordNet is the word "apple". It has two senses as follows:
- apple#n#1: fruit with red or yellow or green skin and sweet to tart crisp whitish flesh.
- apple#n#2: native Eurasian tree widely cultivated in many varieties for its firm rounded edible fruits.

The Wikipedia article "apple" refers to both senses, and so selection of either WordNet sense is correct. It is very difficult for the algorithm to distinguish between them.

### 7.2.2 Disambiguation by mapping the WordNet ontological tree to Wikipedia categories

Our 12,906 articles taken from Wikipedia belong to a total of 18,810 associated categories. Thus, clearly some articles have more than one category; however some articles also do not have any category.
In WordNet there are 107,943 hypernym relations.

|  | Baseline | | Our methods | |
| --- | --- | --- | --- | --- |
|  | VSM | VSM (using first part of the article) | MCAT | VSM+ MCAT |
| Correct sense identification | 410 (81.5%) | 403 (79.48%) | 380 (95%) | **426 (84.02%)** |
| Incorrect sense identification | 97 (18.5%) | 104 (20.52%) | 20 (5%) | 81 (15.98%) |
| Total ambiguous words | 507 (100%) | | 400 (100%) | 507 (100%) |

Table 1. Results of disambiguation

Results:

We successfully aligned 2,239 Wikipedia article titles with a WordNet sense. 400 of the 507 articles in our test data have Wikipedia category pages allowing us apply MCAT. Our human evaluation found that 95% (380 words) were correctly disambiguated. This outperformed disambiguation using VSM, demonstrating the utility of the taxonomic information in Wikipedia and WordNet. However, because not all words in Wikipedia have categories, and there are very few named entities in WordNet, the number of disambiguated words that can be obtained with MCAT (2,239) is less than when using VSM, (12,906).

Using only MCAT reduces the size of the Japanese-Spanish thesaurus. We had the intuition that by combining both disambiguation methods we can achieve a better balance between coverage and accuracy. VSM+MCAT use the MCAT WSD results when available falling back to VSM results otherwise.

We got an accuracy of 84.02% (426 of 507 total words) with VSM+MCAT, outperforming the baselines.

**Evaluating the coverage over Comparable corpus**

- Corpus construction

We construct comparable corpus by extracting from Wikipedia articles content information as follows:
Choose the articles whose content belongs to the thesaurus. We only took the first part of the article until a subtitle and split into sentences.

- Evaluation of coverage

We evaluate the coverage of the thesaurus over an automated comparable corpus automatically extracted from Wikipedia. The comparable corpus consists of a total of 6,165 sentences collected from 12,900 articles of Wikipedia.
We obtained 34,525 types of words; we map them with 15,764 from the Japanese-English-Spanish thesaurus. We found 10,798 types of words that have a coincidence that it is equivalent to 31.27%.

We found this result acceptable for find information inside Wikipedia.

## 8 Conclusion and future work

This paper focused on the creation of a Japanese-Spanish-English thesaurus and ontological relations. We demonstrated the feasibility of using Wikipedia's features for aligning several languages.We present the results of three sub-tasks:

The first sub-task used pattern matching to align the links between Spanish, Japanese, and English articles' titles.

The second sub-task used two methods to disambiguate the English article titles by assigning the WordNet senses to each English word; the first method compares the disambiguation using cosine similarity. The second method uses Wikipedia categories. We established that using Wikipedia categories and the WordNet ontology gives promising results, however the number of words that can be disambiguated with this method is small compared to the VSM method. However, we showed that combining the two methods achieved a favorable balance of coverage and accuracy.

Finally, the third sub-task involved translating English thesaurus entries into Spanish and Japanese to construct a multilingual aligned thesaurus.

So far most of research on Wikipedia focuses on using only a single language. The main contribution of this paper is that by using a huge multilingual data resource (in our case Wikipedia) combined with a structured monolingual resource such as WordNet, we have shown that it is possible to extend a monolingual resource to other languages. Our results show that the method is quite consistent and effective for this task.

The same experiment can be repeated using Wikipedia and WordNet on languages others than Japanese and Spanish offering useful results especially for minority languages.

In addition, the use of Wikipedia and WordNet in combination achieves better results than those that could be achieved using either resource independently.

We plan to extent the coverage of the thesaurus to other syntactic categories such as verbs, adverb, and adjectives. We also evaluate our thesaurus in real world tasks such as the construction of comparable corpora for use in MT.

### Acknowledgments

We would like to thanks to Eric Nichols for his helpful comments.